\pdfoutput=1

\documentclass[11pt]{article}

\usepackage{emnlp2021}

\usepackage{times}
\usepackage{latexsym}

\usepackage[T1]{fontenc}

\usepackage[utf8]{inputenc}

\usepackage{microtype}
\usepackage{float}
\usepackage{array}
\usepackage{mathtools}
\usepackage[export]{adjustbox}
\usepackage{xcolor}
\usepackage{graphicx}
\usepackage{natbib}
\usepackage{enumitem}
\usepackage{booktabs}
\usepackage{amsmath}
\usepackage{multirow}
\usepackage{arydshln}
\usepackage{siunitx}
\usepackage{enumitem}

\usepackage{subcaption}

\graphicspath{ {./images/} }

\newcommand{\bertbase}{BERT$_{\textsc{base}}$ }
\newcommand{\bertlarge}{BERT$_{\textsc{large}}$ }
\newcommand{\robertabase}{RoBERTa$_{\textsc{base}}$ }
\newcommand{\camembase}{CamemBERT$_{\textsc{base}}$ }

\title{How to Train BERT with an Academic Budget} 

\author{Peter Izsak$^\clubsuit$ \quad Moshe Berchansky$^\clubsuit$ \quad Omer Levy$^\diamondsuit$ \\
$^{\clubsuit}$Intel Labs, Israel \\
$^\diamondsuit$Blavatnik School of Computer Science, Tel Aviv University \\
{\tt \{peter.izsak,moshe.berchansky\}@intel.com} \\[4pt]
{\tt levyomer@cs.tau.ac.il} \\[4pt]
}

\begin{document}

\maketitle

\begin{abstract}
While large language models à la BERT are used ubiquitously in NLP, pretraining them is considered a luxury that only a few well-funded industry labs can afford.
How can one train such models with a more modest budget?
We present a recipe for pretraining a masked language model in 24 hours using a single low-end deep learning server.
We demonstrate that through a combination of software optimizations, design choices, and hyperparameter tuning, it is possible to produce models that are competitive with \bertbase on GLUE tasks at a fraction of the original pretraining cost.\footnote{Our code is publicly available at: \url{https://github.com/IntelLabs/academic-budget-bert}}

\end{abstract}
\section{Introduction} \label{sec:intro}

Large language models, such as BERT~\cite{devlin-etal-2019-bert}, RoBERTa~\cite{Liu2019RoBERTaAR}, and GPT3 \cite{gpt3}, have become the \textit{de facto} models used in many NLP tasks. 
However, their pretraining phase can be prohibitively expensive for startups and academic research groups, limiting the research and development of model pretraining to only a few well-funded industry labs.
How can one train a large language model with commonly-available hardware in reasonable time?

We present a recipe for training a BERT-like masked language model (MLM) in 24 hours in a limited computation environment.
Our approach combines multiple elements from recent work: faster implementation \cite{deepspeed}, faster convergence through over-parameterization \cite{Li2020TrainLT}, best practices for scaling language models \cite{Kaplan2020ScalingLF}, single-sequence training \cite{joshi-etal-2020-spanbert, Liu2019RoBERTaAR},
and more.
Moreover, we conduct an extensive hyperparameter search tailored to our resource budget, and find that synchronizing learning rate warmup and decay schedules with our 24 hour budget greatly improves model performance.

When evaluating on GLUE \cite{wang-etal-2018-glue}, our recipe produces models that are competitive with \bertbase{} -- a model that was trained on 16 TPUs for 4 days. 
This recipe can also be applied to other corpora, as we demonstrate by training a French-language model on par with \camembase \cite{martin-etal-2020-camembert} on the XNLI French benchmark \cite{conneau2018xnli}.
Overall, our findings demonstrate that, with the right recipe and an understanding of the available computational resources, large language models can indeed be trained in an academic setting.

\section{Problem Setup}
\label{sec:problem}

We investigate the task of pretraining a large language model under computational constraints.
To simulate an academic computation budget, we limit the training time to 24 hours and the hardware to a single low-end deep learning server.\footnote{Specifically, we experiment with 8 Nvidia Titan-V GPUs with 12GB memory each. In terms of GB-hour, our setting is roughly equivalent to 1 day with 4 RTX 3090 GPUs or 2.4 days on a single 40GB A100 GPU.}
Using current cloud-compute prices, we estimate the dollar-cost of each training run at around \$50 to \$100.

Under these constraints, our goal is to pretrain a model that can benefit \textit{classification} tasks, such as in GLUE \cite{wang-etal-2018-glue}.
Therefore, we follow the standard practice and focus on BERT-style transformer encoders trained on the MLM objective \cite{devlin-etal-2019-bert}.
We retain the standard pretraining corpus of English Wikipedia and the Toronto BookCorpus \cite{zhu2015aligning}, containing 16GB of text, tokenized into subwords using BERT's uncased tokenizer.

\section{Combining Efficient Training Methods}
\label{sec:methods}

To speed up our training process, we combine a variety of recent techniques for optimizing a masked language model.
To the best of our knowledge, this is the first time that such techniques are combined and evaluated as a unified framework for training large models with limited computational resources.


\subsection{Methods}
\label{sec:methods-methods}

\paragraph{Data}
Since our focus is mainly on sentence classification tasks, we limit sequences to 128 tokens for the entire pretraining process.
\citet{devlin-etal-2019-bert} also apply this practice to 90\% of the training steps, and extend the sequence to 512 tokens for the last 10\%.
This increases sample efficiency by reducing padding, and also allows us to fit a larger model into memory (see \textit{Model}).
In addition, we use single-sequence training without the next sentence prediction (NSP) objective, which was shown to benefit optimization \cite{joshi-etal-2020-spanbert, Liu2019RoBERTaAR}.
To maximize time spent on training, we hold out only 0.5\% of the data and compute the validation-set loss every 30 minutes.


\paragraph{Model}
Recent work has found that larger models tend to achieve better performance than smaller models when trained for the same wall-clock time \cite{Li2020TrainLT, Kaplan2020ScalingLF}.
We adopt these recommendations and train a \bertlarge model: 24 layers, 1,024 dimensions, 16 heads, 4,096 hidden dimensions in the feed-forward layer, with pre-layer normalization \cite{Shoeybi2019MegatronLMTM}.
The purpose of applying the ``train large'' approach is \textit{not} to compete with fully-trained extra-large models, but to train the best model we can, regardless of size, given the computational constraints  (Section~\ref{sec:problem}).

\paragraph{Optimizer}
We follow the optimization of RoBERTa \cite{Liu2019RoBERTaAR} and use AdamW \cite{Loshchilov2019DecoupledWD} with $\beta_1 = $~0.9, $\beta_2 = $~0.98, $\varepsilon = $~1e-6, weight decay of 0.01, dropout 0.1, and attention dropout 0.1.
We experiment with various learning rates and warmups in Section~\ref{sec:hyperparameters}. 
Preliminary experiments with other optimizers, such as LAMB~\cite{you2020large}, did not yield significantly different trends.

\paragraph{Software}
We base our implementation on the DeepSpeed software package \cite{deepspeed}, which includes optimizations for training language models, such as data parallelization, and mixed-precision training.
We further improve the implementation by replacing the MLM prediction head with sparse token prediction \cite{Liu2019RoBERTaAR}, and use fused implementations for all linear-activation-bias operations and layer norms, in particular the APEX LayerNorm operation.

\paragraph{I/O}
To reduce the I/O bottleneck and minimize time wasted on non-training operations, we follow \citet{devlin-etal-2019-bert} and pre-mask 10 copies of the corpus.
While \citet{Liu2019RoBERTaAR} recommends dynamic masking, the benefits of applying it in our low-resource setting are marginal, and outweighed by the computational cost.
To ensure heterogeneous mini-batches, we shuffle the entire dataset after masking to remove intra-shard duplicates.
Finally, we avoid disk I/O by sharding offline and loading the entire preprocessed dataset into RAM.

\subsection{Combined Speedup}

We compare our optimized framework to the official implementation of \citet{devlin-etal-2019-bert}.\footnote{\url{https://github.com/tensorflow/models/tree/master/official/nlp/bert}}
Table~\ref{tab:backend} shows that using the official code to train \bertbase could take almost \textit{6 days} under our hardware assumptions (Section~\ref{sec:problem}), and a large model might require close to \textit{a month} of non-stop computation.
In contrast, our recipe significantly speeds up training, allowing one to train \bertlarge with the original number of steps (1M) in a third of the time (8 days), or converge in 2-3 days by enlarging the batch size.
While larger batch sizes do not guarantee convergence to models of equal quality, they are generally recommended \cite{ott-etal-2018-scaling,Liu2019RoBERTaAR}, and present a more realistic starting point for our next phase (hyperparameter tuning) given our 24-hour constraint.

\begin{table}
\centering
\small
\begin{tabular}{@{}llccc@{}}
\toprule
 & \textbf{~~~~bsz} & \textbf{steps} & \textbf{samples} & \textbf{days} \\ \midrule
Google \bertbase                   & ~~~~256   & 1000k    & 256M  & ~~5.85  \\ 
Google \bertlarge        & ~~~~128$^\dagger$    & 2000k    & 256M   & 26.33 \\ \midrule 
\multirow{5}{*}{Our \bertlarge{}}    & ~~~~128   & 2000k    & 256M     & 14.11 \\
    & ~~~~256   & 1000k    & 256M     & ~~8.34 \\
    & ~~4096     & ~~~~63k & 256M    & ~~2.74 \\
    & ~~8192     & ~~~~31k & 256M     & ~~2.53 \\
    & 16384      & ~~~~16k & 256M & ~~2.41   \\ \bottomrule
\end{tabular}
\caption{Speed comparison between our optimized framework and the official implementation of BERT, while testing on the same hardware and controlling for the number of training examples covered (256M).
$^\dagger$Largest batch size we could fit (128), requiring double the steps to cover the same amount of examples.
}
\label{tab:backend}
\end{table}


We also conduct an ablation study of engineering improvements in our model.
Table~\ref{tab:ablation_study} shows that efficient implementation gains an additional 1.75 hours (out of 24) for training operations, which would have otherwise been wasted.
 
\begin{table}[t]
\centering
\small
\begin{tabular}{@{}lr@{}}
\toprule
&
\bf Time Saved \\
\midrule
$-$ Sparse Output Prediction & -3.91\%  \\
$-$ Fused Linear Layer & -4.39\%  \\
$-$ APEX LayerNorm & -7.28\% \\
\bottomrule
\end{tabular}
\caption{Speed up (in cumulative training time reduction) for each implementation improvement in our framework. Each line represents the original model without the measured feature, aggregated with preceding feature.}
\label{tab:ablation_study}
\end{table}

\begin{figure}[ht]
    \centering
    \begin{subfigure}[b]{0.2\textwidth}
        \includegraphics[width=\textwidth]{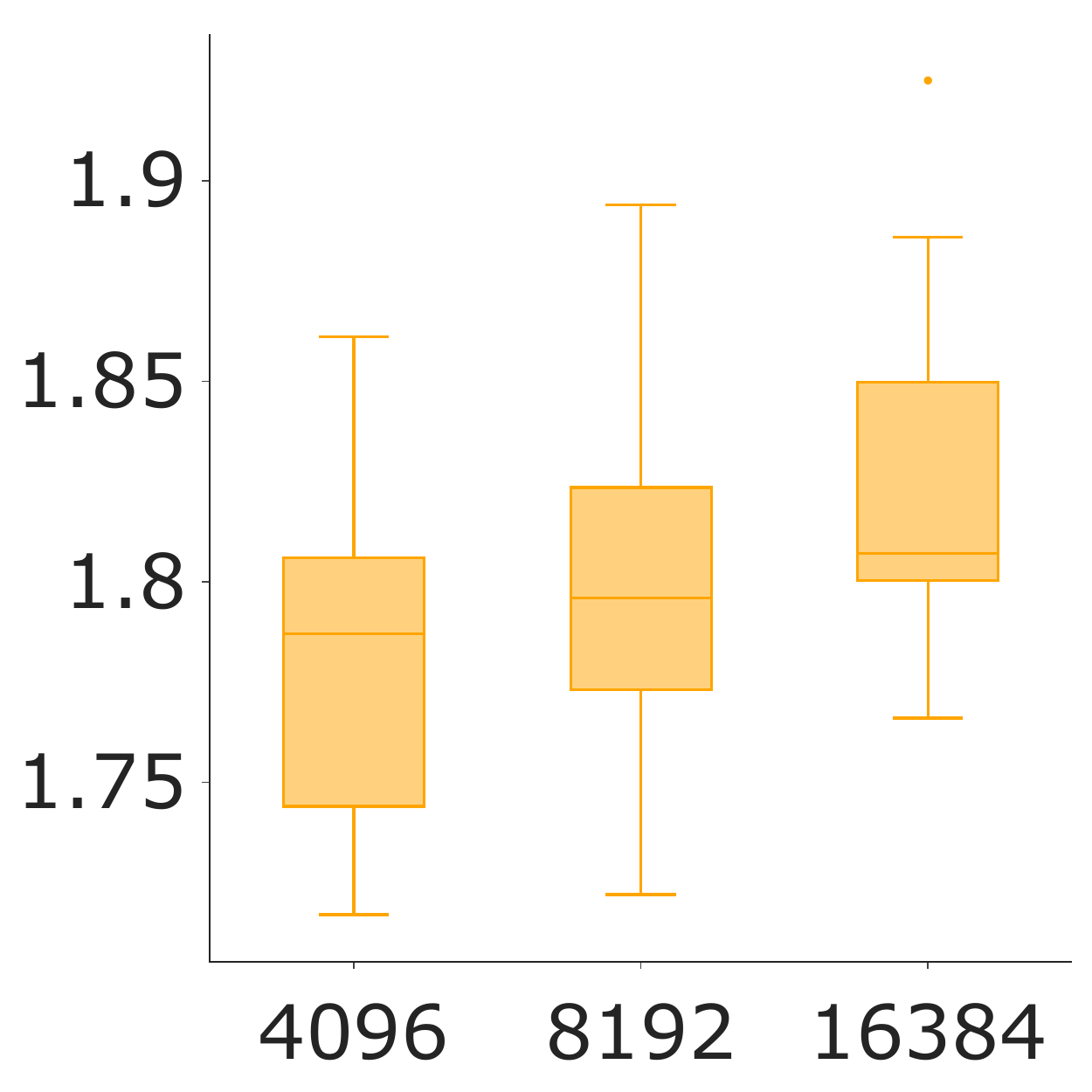}
        \caption{Batch Size}
        \label{fig:bs_dist}
    \end{subfigure}
    ~~~~~~
    \begin{subfigure}[b]{0.2\textwidth}
        \includegraphics[width=\textwidth]{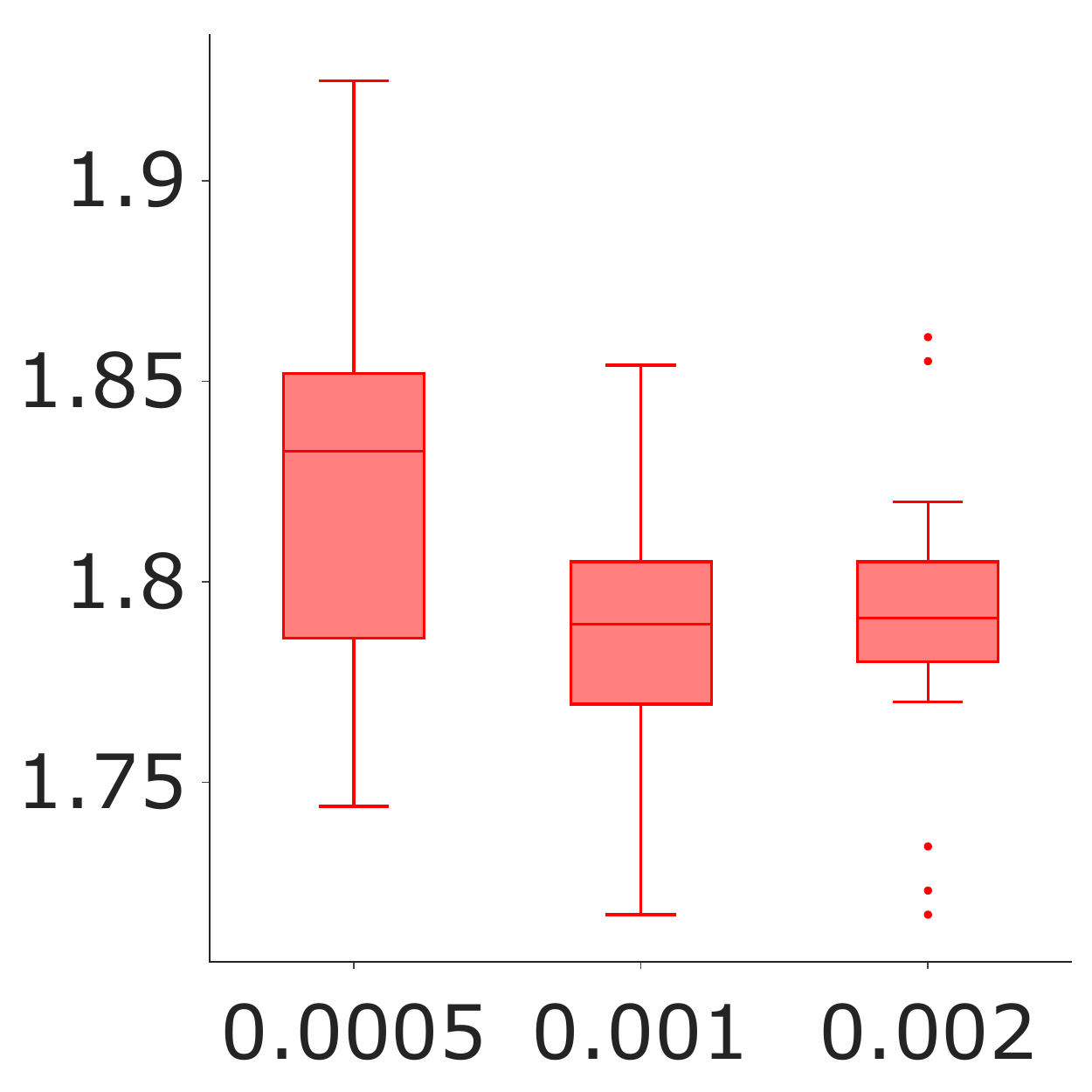}
        \caption{Learning Rate}
        \label{fig:lr_dist}
    \end{subfigure}
    \\
    \begin{subfigure}[b]{0.2\textwidth}
        \includegraphics[width=\textwidth]{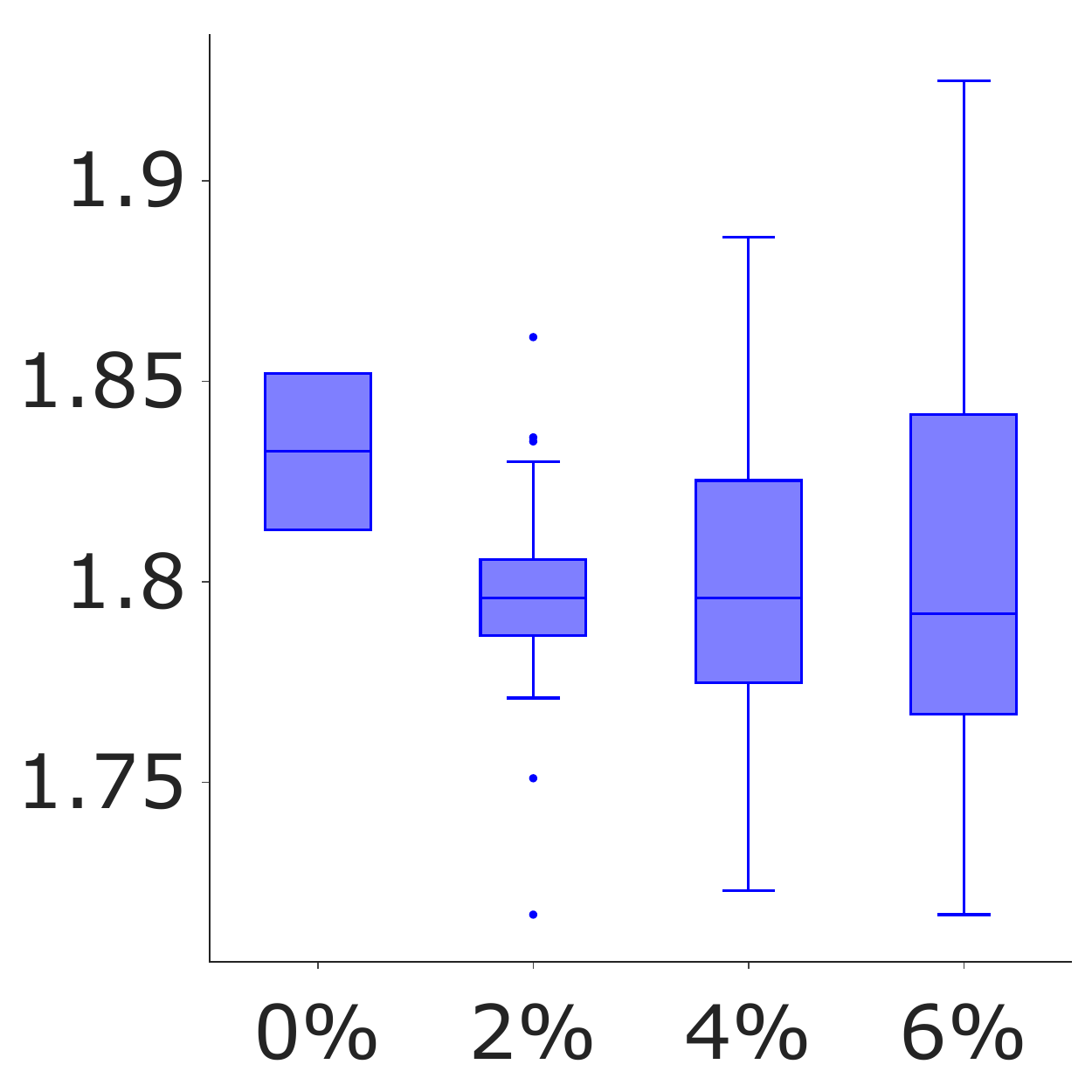}
        \caption{Warmup Proportion}
        \label{fig:warmup_proportion_dist}
    \end{subfigure}
    ~~~~~~
    \begin{subfigure}[b]{0.2\textwidth}
        \includegraphics[width=\textwidth]{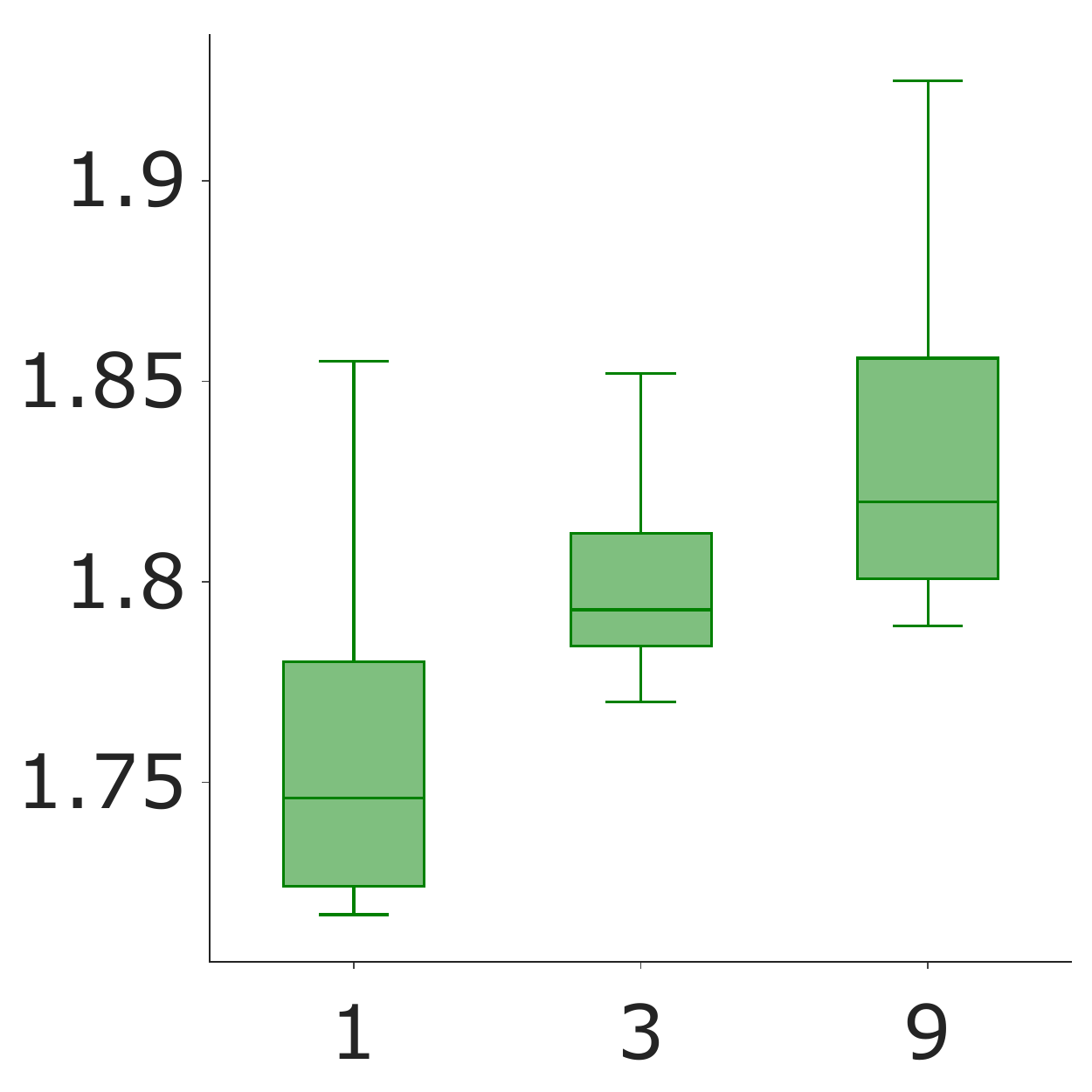}
        \caption{Total Days}
        \label{fig:total_day_dist}
    \end{subfigure}
    \caption{Distribution of the validation-set loss after 24 hours of training across different hyperparameters.}
    \label{fig:24_hour_sweep}
\end{figure}

\section{Hyperparameter Search}
\label{sec:hyperparameters}

Calibrating hyperparameters is key to increasing model performance in deep learning and NLP \cite{levy-etal-2015-improving, Liu2019RoBERTaAR}.
We re-tune core optimization hyperparameters to fit our low-resource setting, rather than the massive computation frameworks for which they are currently tuned.
Our hyperparameter search yields substantial improvements in MLM loss after 24 hours of training.

\subsection{Hyperparameters}
\label{subsection:hyperparams}


\paragraph{Batch Size (bsz)}
The number of examples (sequences up to 128 tokens) in each mini-batch.
We try batch sizes of 4k, 8k, and 16k examples, which are of a similar order of magnitude to the ones used by \citet{Liu2019RoBERTaAR}.
Since our hardware has limited memory, we achieve these batch sizes via gradient accumulation.
In terms of parameter updates, these batch sizes amount to approximately 23k, 12k, and 6k update steps in 24 hours, respectively.

\paragraph{Peak Learning Rate (lr)}
Our linear learning rate scheduler, which starts at 0, warms up to the peak learning rate, and then decays back to 0. We try 5e-4, 1e-3, and 2e-3.

\paragraph{Warmup Proportion (wu)}
We determine the number of warmup steps as a proportion of the total number of steps. Specifically, we try 0\%, 2\%, 4\%, and 6\%, which all reflect significantly fewer warmup steps than in BERT.

\paragraph{Total Days (days)}
The number of days it would take the learning rate scheduler to decay back to 0, as measured on our hardware. This is equivalent to setting the maximal number of steps.
Together with the warmup proportion, it determines where along the learning rate schedule the training process stops.
For a value of 1 day, the learning process will end when the learning rate decays back to 0.
We try setting the schedule according to 1, 3, and 9 days.

\subsection{Methodology}

We optimize our model using MLM loss with each hyperparameter setting.
Although there are 108 combinations in total, poor configurations are easy to identify early on.
After 3 hours, we prune configurations that did not reach a validation-set loss of 6.0 or less; this rule removes diverging runs, such as configurations with 0\% warmup.
After 12 hours, we keep the top 50\% of models with respect to the validation-set loss, and resume their runs until they reach 24 hours.

\begin{table}[t]
\centering
\small
\begin{tabular}{@{}lccccc@{}}
\toprule
\textbf{Configuration} &
\bf loss &
\bf bsz &
\bf lr &
\bf wu &
\bf days \\
\midrule
Search \#1 & 1.717 & 4096 & 2e-3 & 6\% & 1 \\
Search \#2 & 1.717 & 4096 & 1e-3 & 2\% & 1 \\
Search \#3 & 1.720 & 4096 & 1e-3 & 6\% & 1 \\
Search \#4 & 1.722 & 8192 & 1e-3 & 6\% & 1 \\
Search \#5 & 1.723 & 4096 & 2e-3 & 4\% & 1 \\
\midrule
\bertbase{}   & 2.050 & 256 & 1e-4 & 11.1\% & 3 \\
\bertlarge{}  & 2.318 & 256 & 1e-4 & 11.1\% & 8.3 \\
\bottomrule
\end{tabular}
\caption{Best hyperparameter configurations by MLM loss recorded after 24 hours of training.}
\label{tab:best_24hr_configs}
\end{table}

\begin{figure}
\centering
\includegraphics[width=\columnwidth]{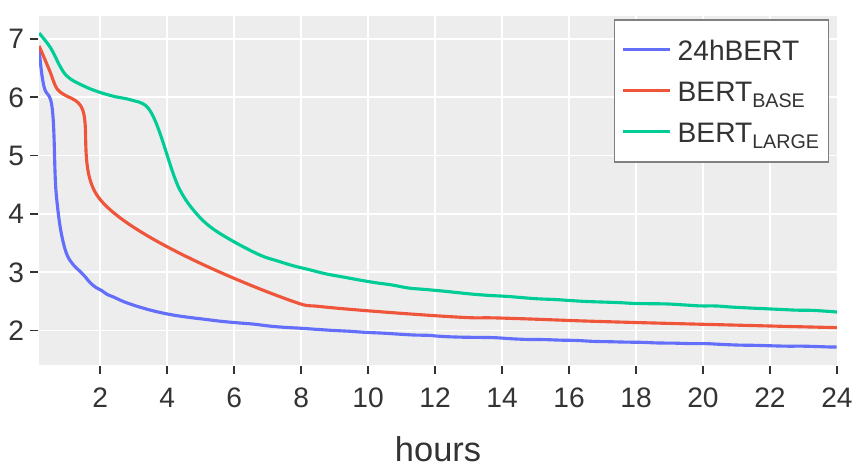}
\caption{The validation-set loss of 24hBERT compared to the original BERT model configurations.}
\label{fig:bert_comparison_in_24hr}
\end{figure}

\subsection{Results}

\begin{table*}[!ht]
\centering
\small
\begin{tabular}{@{}lccccccccc@{}}
\toprule
& \bf MNLI-m/mm & \bf QNLI & \bf QQP & \bf RTE & \bf SST-2 & \bf MRPC & \bf CoLA & \bf STS-B & \bf Avg. \\
\#Examples & 393k & 105k & 364k & 2.5k & 67k & 3.7k & 8.5k & 7k &  \\
\midrule 


\textbf{24hBERT}    & 84.4/83.8 & 90.6 & 70.7 & 75.3 & 93.0 & 88.5 & 57.1 & 86.8 & 81.1 \\
\bertbase           & 84.6/84.0 & 90.6 & 72.0 & 76.5 & 92.8 & 89.9 & 55.1 & 87.7 & 81.5 \\
\bertlarge          & 86.0/85.2 & 92.6 & 72.0 & 78.3 & 94.5 & 89.9 & 60.9 & 87.5 & 83.0 \\
\robertabase        & 87.0/86.5 & 92.4 & 72.5 & 79.6 & 95.8 & 89.7 & 58.8 & 88.3 & 83.4 \\

\bottomrule
\end{tabular}
\caption{Performance on GLUE test sets. Results for RTE, STS and
MRPC are reported by first finetuning on the MNLI model instead of the baseline pretrained model.
}

\label{tab:glue_tasks}
\end{table*}

We first analyze the effect of each hyperparameter by plotting the distribution of the validation-set loss per value (Figure~\ref{fig:24_hour_sweep}).
We observe a clear preference towards synchronizing the learning rate schedule with the actual amount of training time in the budget (1 day), corroborating the results of \citet{Li2020BudgetedTR}.
We also find the smaller batch size to have an advantage over larger ones, along with moderate-high learning rates.
We suspect that the smaller batch size works better for our resource budget due to the trade-off between number of samples and number of updates, for which a batch size of 4096 seems to be a better fit.
Finally, there appears to be a preference towards longer warmup proportions; however, a closer look at those cases reveals that when the number of total days is larger (3 or 9), it is better to use a smaller warmup proportion (2\%), otherwise the warmup phase might take up a larger portion of the \textit{actual} training time.

Table~\ref{tab:best_24hr_configs} shows the best configurations by MLM
loss.
It is apparent that our calibrated models perform substantially better than models with BERT’s default hyperparameters (which were tuned for 4 days on 16 TPUs). 
There is also relatively little
variance in performance among the top models.
We select the best model (Search \#1), and name it \textbf{24hBERT}.
Figure~\ref{fig:bert_comparison_in_24hr} compares 24hBERT with models using the default calibration, and shows that 24hBERT converges significantly faster. 






\section{Downstream Evaluation}
\label{sec:tasks_eval}

We test the performance of our optimized, calibrated 24hBERT model on the GLUE benchmark~\cite{wang-etal-2018-glue}.\footnote{See Appendix~\ref{appendix:c} for a full description of tasks.}
For finetuning, we follow the practice of \citet{Liu2019RoBERTaAR}, and run a grid search over multiple hyperparameters and seeds (see Appendix~\ref{appendix:b}), and also use mid-training~\cite{phang2018stilts} on MNLI for RTE, MRPC and STS-B.

Table~\ref{tab:glue_tasks} shows the results on GLUE's test sets.
Our 24hBERT model performs on par with \bertbase on 3 major tasks (MNLI, QNLI, SST-2) and even outperforms it on CoLA. However, 24hBERT reaches slightly lower results on 4 tasks (QQP, RTE, MRPC, STS-B). Overall, this amounts to a small difference on the average score (0.4\%), showing that our recipe can indeed produce a model that is largely competitive with \bertbase, but at a small fraction of its training cost.






\section{Generalizing to New Corpora}

Our recipe was calibrated using a particular corpus (English Wikipedia and books), but does it generalize to other corpora as well?
We follow CamemBERT \cite{martin-etal-2020-camembert} and train a masked language model on French Wikipedia, using exactly the same dataset.
We then finetune our French 24hBERT on the XNLI French dataset \cite{conneau2018xnli}, reaching 78.5\% accuracy, compared to  79.1\% of \camembase{}.
This result demonstrates that our recipe can indeed be ported to other corpora as-is, without retuning hyperparameters.

\section{Discussion}

\begin{figure}
\centering
\includegraphics[width=\columnwidth]{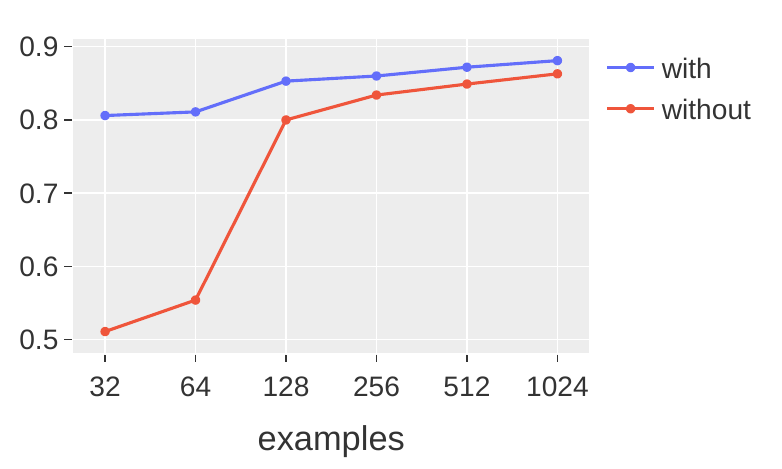}
\caption{The performance of 24hBERT on the SST-2 task in few-shot settings (5 seeds for each \#examples, with 25\% of the examples used for validation), with and without prompts.}
\label{fig:prompts_eval}
\end{figure}

\paragraph{Comparison with ELECTRA}
While \citet{clark-etal-2020-pre} show impressive pretraining speedups with ELECTRA, we argue that having a generative model (MLM or LM) is important nowadays, given the recent rise of few-shot learning and prompting approaches \cite{schick-schutze-2021-exploiting}.
To emphasize this point, we run 24hBERT on the SST-2~\cite{socher-etal-2013-recursive} task both with and without prompts in the few-shot setting.
Figure~\ref{fig:prompts_eval} shows that there is a significant advantage in the ability to prompt the model, which is perhaps not trivial for non-generative ELECTRA-style models.



\paragraph{FLOPs as a Measure of Efficiency}
While measuring floating point operations is commonly used to compare efficiency in a hardware-agnostic manner, it is \textit{not} an accurate tool for comparing the actual time (and therefore budget) associated with training a model.
Specifically, measuring FLOPs ignores the fact that many operations run in parallel (e.g. via batching), and are thus much less costly in practice \cite{Li2020TrainLT}.




\paragraph{Limitations}
Our investigation is limited to classification tasks.
While it is true that it is not fully comparable with \bertbase in that using short sequences does not allow for reading comprehension tasks (without resorting to sliding windows), it might be possible to continue training the model for a few more hours with sequences longer than 128 tokens, as done by \citet{devlin-etal-2019-bert}.
We leave such experiments for future work.


\section{Conclusions}

We present a recipe for pretraining a masked language model in 24 hours using a low-end deep learning server.
We show that by combining multiple efficient training methods presented in recent work and carefully calibrating the hyperparameters it is possible to pretrain a model that is competitive to \bertbase on GLUE tasks.
In contrast to other works in this area, which often focus a single method for improving efficiency, our recipe consists of many different components that together amount to very large speedups:
\begin{itemize}[itemsep=0pt,topsep=3pt,leftmargin=10pt]
    \item Short sequences~\cite{devlin-etal-2019-bert}
    \item Single-sequence training~\cite{joshi-etal-2020-spanbert}
    \item Training larger models~\cite{Li2020TrainLT}
    \item DeepSpeed~\cite{deepspeed}
    \item Sparse token prediction~\cite{Liu2019RoBERTaAR}
    \item Fused implementations
    \item Avoiding disk I/O
    \item Large batch sizes~\cite{Liu2019RoBERTaAR}
    \item Large learning rates~\cite{Liu2019RoBERTaAR}
    \item Short warmup
    \item Synchronizing schedule with time budget~\cite{Li2020BudgetedTR}
\end{itemize}
As with every recipe, our recommendations may need to be adapted to the hardware and time constraints at hand.
We hope that our findings allow additional players to participate in language model research and development, and help democratize the art of pretraining.


\bibliographystyle{acl_natbib}
\bibliography{acl_anthology,references}

\newpage

~

\newpage
\appendix




\section{Pretraining Hyperparameters}
\label{appendix:a}

Table~\ref{tab:pretrain_hp} presents the full set of hyperparameter configurations we examine in Section~\ref{sec:hyperparameters}.

\section{Finetuning Hyperparameters} 
\label{appendix:b}

Finetuning hyperparameters used for the GLUE benchmark tasks are presented in Table~\ref{tab:glue_hp}. We run each configuration using 5 random seeds and select the median of the best configuration.

\section{Performance Comparison}

Table~\ref{tab:backend_others} includes time comparison of our 24 hour training setup when using more recent hardware backends.

\section{Downstream Tasks}
\label{appendix:c}


\textbf{MNLI}: Multi-Genre Natural Language Inference
is a large-scale, crowd-sourced entailment classification task \cite{williams2018broadcoverage}. Given a pair of
sentences, we wish to predict whether the second sentence is an entailment, contradiction, or neutral with respect to the first one.

\textbf{QQP}: Quora Question Pairs is a binary classification task, where the goal is to determine whether two
questions asked on Quora are semantically equivalent or not \cite{qqpcite}.

\textbf{QNLI}: Question Natural Language Inference is
a version of the Stanford Question Answering
Dataset \cite{rajpurkar2016squad}. It has been
converted into a binary classification task \cite{wang-etal-2018-glue}. The positive examples are (question, sentence) pairs, which contain the answer, and the negative examples are from the same paragraph, yet do not contain the answer.

\textbf{SST-2}: The Stanford Sentiment Treebank is a
binary single-sentence classification task, consisting of sentences extracted from movie reviews. Their sentiment is based on human annotations \cite{socher-etal-2013-recursive}.

\textbf{CoLA}: The Corpus of Linguistic Acceptability is
a binary single-sentence classification task, where
the goal is to predict whether an English sentence
is linguistically “acceptable” or not \cite{warstadt-etal-2019-neural}.

\textbf{STS-B}: The Semantic Textual Similarity Benchmark is a collection of sentence pairs, drawn primarily from news headlines, with additional sources as well \cite{cer-etal-2017-semeval}. They were annotated with a score from 1 to 5, which denotes how similar the two sentences are, when semantic meaning is considered.

\textbf{MRPC}: Microsoft Research Paraphrase Corpus
consists of sentence pairs automatically extracted
from online news sources. The human annotations are
for whether the sentences in the pair are semantically equivalent \cite{dolan-brockett-2005-automatically}.

\textbf{RTE}: Recognizing Textual Entailment is a binary entailment task similar to MNLI, but with significantly less training data \cite{10.1007/11736790_9, bar-haim-2006, 10.5555/1654536.1654538}.

\textbf{XNLI French}: Cross-lingual Natural Language Inference French, an entailment classification task \cite{conneau2018xnli} similar to MNLI, with that the premise and hypothesis in each example are in the French language.

\begin{table}[t!]
\centering
\small
\begin{tabular}{@{}lcc@{}}
\toprule
\bf Hyperparameter  & \bf Our Model \\
\midrule 
Number of Layers & 24 \\
Hidden size & 1024 \\
FFN inner hidden size & 4096 \\
Attention heads & 16 \\
Attention head size & 64 \\
Dropout & 0.1 \\
Attention Dropout & 0.1 \\
Learning Rate Decay & Linear \\
Weight Decay & 0.01 \\
Optimizer & AdamW \\
Adam $\epsilon$ & 1e-6 \\
Adam $\beta_1$ & 0.9 \\
Adam $\beta_2$ & 0.98 \\
Gradient Clipping & 0.0 \\ \midrule
Batch Size & \{4096, 8192, 16384\} \\
Peak Learning Rate & \{5e-4, 1e-3, 2e-3\} \\
Warmup Proportion & \{0\%, 2\%, 4\%, 6\%\} \\
Max Steps & \{24hr, 72hr, 216hr\} \\
\bottomrule
\end{tabular}
\caption{
Hyperparameters used for pretraining our models.
}
\label{tab:pretrain_hp}
\end{table}

\begin{table}[t]
    \small
    \centering
    \begin{tabular}{@{}lcrrr@{}}
    \toprule
    & \textbf{GPUs} & \textbf{Days} & \textbf{BSZ/GPU} & \textbf{ACC} \\
    \midrule
    Titan-V 12GB & 8 & 1.00 & 32 & 16 \\
    \midrule
    \multirow{2}{*}{RTX 3090 24GB} & 1 & 5.84 & 112 & 37 \\
    & 4 & 1.55 & 112 & 9 \\
    \midrule
    \multirow{2}{*}{A100 40GB} & 1 & 2.75 & 200 & 20 \\
     & 4 & 0.74 & 200 & 5 \\
    \bottomrule 
    \end{tabular}
    \caption{Number of days, batch size per GPU (BSZ/GPU), and number of gradient accumulations (ACC) to train a model using our recipe (24hBERT) with more recent GPUs.}
    \label{tab:backend_others}
\end{table}
\begin{table*}[t]
\centering
\small
\begin{tabular}{@{}lccc@{}}
\toprule
\bf Hyperparameter  & \bf RTE, SST-2, MRPC, CoLA, STS-B, WNLI & \bf MNLI, QQP, QNLI, XNLI French \\
\midrule 
Learning Rate & \{1e-5, 3e-5, 5e-5, 8e-5\} & \{5e-5, 8e-5\}\\
Batch Size & \{16, 32\} & 32\\
Weight Decay & 0.1 & 0.1 \\
Max Epochs & \{3, 5, 10\} & \{3, 5\} \\
Warmup Proportion & 0.06 & 0.06 \\
\bottomrule
\end{tabular}
\caption{
The hyperparameter space used for finetuning our model on GLUE benchmark tasks, and the XNLI French task.
}
\label{tab:glue_hp}
\end{table*}



\end{document}